# Symbol Grounding in Computational Systems: A Paradox of Intentions

Vincent C. Müller

Anatolia College/ACT, P.O. Box 21021, 55510 Pylaia, Greece
vmueller@act.edu, http://www.typos.de

**Abstract.** The paper presents a paradoxical feature of computational systems that suggests that computationalism cannot explain symbol grounding. If the mind is a digital computer, as computationalism claims, then it can be computing either over meaningful symbols or over meaningless symbols. If it is computing over meaningful symbols its functioning presupposes the existence of meaningful symbols in the system, i.e. it implies semantic nativism. If the mind is computing over meaningless symbols, no intentional cognitive processes are available prior to symbol grounding; therefore no symbol grounding could take place since any such process presupposes intentional processes. So, whether computing in the mind is over meaningless or over meaningful symbols, computationalism implies semantic nativism.

## 1. Computationalism

We will discuss an aspect of the problems a digital computational system has to acquire meaningful symbols and what these problems mean for a computational theory of the mind, in particular its relation to semantic nativism. The classical computational theory of the mind, or "computationalism" for short, holds that the mind is a digital computer, in particular that it is a computational information processor. The view of the mind as a computer, especially as a computer processing symbols according to rules, is the basis of classical cognitive science and artificial intelligence. As Fodor puts it: "The cognitive science that started fifty years or so ago more or less explicitly had as its defining project to examine a theory, largely owing to Turing, that cognitive mental processes are operations defined on syntactically structured mental representations that are much like sentences." (Fodor 2000: 3-4).

Computationalism is typically (but not necessarily) a version of the view of mental states as physical states with a specific causal *functional role*, as proposed by the earlier Putnam. If the mind is described not at the a basic physical level, but described at the level of these functional roles and if these are taken as realizations of a Turing machine, as computational states, then we have the theory commonly known as *Machine Functionalism*, which includes the thesis of the necessity of computing for mentality: "Mentality, or having a mind, consists in realizing an appropriate Turing machine" (Fodor 1994: 10-15; Kim 1996: 91)—a thesis that is stronger than computationalism itself. Paul Churchland characterizes the thesis as follows: "What unites them [the cognitive creatures] is that (…) they are all computing the same, or some part of the same abstract <<sensory input, prior state>, <motor output, subsequent state>> *function*." (Churchland 2005: 333). The computationalist version of functionalism is initially plausible because computers are necessarily described functionally, as in the notion of a "Turing machine." It does not make sense to describe the mind as a computer in the sense of an identity theory because the *physical description* of a particular computing machine is irrelevant, what matters is the *syntactical description* of its function, and there could be well be such a description of a brain (since nobody claims that our brain physically consists of silicon chips like the ones used in our PCs).

We shall only discuss computationalism in the sense that computation is sufficient for mental states and that it is the cause of mental states in humans, not in the stronger sense that computation is necessary and sufficient (or only necessary). It must be noted, however, that computationalism is not just the weak thesis that some or all mental processes can be *modeled* on a digital computer. If a hurricane can be modeled on a computer, this is not to say that the hurricane *is* a computational system. (NB, it is doubtful whether such modeling is strictly speaking possible on a digital computer, since a hurricane is not a discrete state phenomenon). Despite the distinction of computationalism from this weaker thesis, there is the possibility, however, that minds might be special cases such that modeling a mind actually *is* producing a mind—given that it has sufficient functional properties (e.g. Chalmers 1996: 328).

Computationalism directly implies the possibility of strong Artificial Intelligence: "… computers can think because, in principle, they can be programmed with the same program that constitutes human thought." (Wakefield 2003: 286). Or, as Churchland puts it: "The central job of AI research is to create *novel physical realizations* of salient parts of, and ultimately all of, the abstract function we are all (more or less) computing." (Churchland 2005: 34).

The notion of computing used her is the classical one as defined by Turing, i.e. *computing* means a mechanical procedure (e.g. the manipulation of symbols) according to *algorithms*, i.e. explicit non-ambiguous rules that proceed step by step and that can be carried out in finite time, leading to a definite output—what is also called "effective computing." The Church-Turing thesis says that a Turing machine can compute all and only the effectively computable functions. (I ignore the theoretical possibility of hypercomputing in this paper, but see (Müller 2006).)

## 2. Computing with Meaningful Symbols: Language of Thought Computationalism

The main theoretical options within computationalism depend on whether the symbols on which the computer operates (and that constitute its program) are meaningful or not. I shall call the option of operating on *meaningful* symbols "Language of Thought Computationalism" or LOCO.

The tradition of Fodor's "Language of Thought" focuses on "cognition" or, even more narrowly, "thought", and it claims that thinking is computing over mental representations. Fodor's slogan could be said to be "There is no computation without representation" (Fodor 1981: 180), so the computing is computing over symbols that represent.

The symbols are also taken to be closely related to natural language concepts, in what Smolensky calls the "Newell/Simon/Fodor/Pylyshyn view": the programs of this computational system "are composed of elements, that is, symbols, referring to essentially the same concepts as the ones used to consciously conceptualize the task domain." (Smolensky 1988: 5; cf. Smolensky 1994). One consequence of this approach is "the view that propositional attitudes (such as believing, noticing, preferring) are to be regarded as computational relations to semantically valuable representations that are encoded in the brain or other hardware of the thinker." (Rey 2002: 203). To conclude in Fodor's words: "The emphasis upon a syntactical character of thought suggests a view of cognitive processes in general—including, for example, perception, memory and learning—as occurring in a languagelike medium, a sort of 'language of thought'." (Fodor 1994: 9). So, LOCO could be summarized as the conjunction of two views:

(1) "Thinking is computation." (Fodor 1998: 9) and
(2) Thinking computes over language-like mental representations.

Fodor's emphasis on the syntactical nature of the computational process should not be taken to mean that his position is anything other than language of thought computationalism. It just so happens, that thinking is a computation over symbols that are representations:

> "First, all mental processes are supposed to be causally sensitive to, and only to, the *syntax* of the mental representations that they are defined over; in particular, mental processes aren't sensitive to what mental representations mean. This is, I think, at the very heart of the Classical [Fodor's] account of cognition." (Fodor 2005: 26)

Given that we have explained the central term of the first thesis (computing), it remains to specify what we mean that of the second: "language." I will just adopt the proposal by Lycan, who says: "(1) they are composed of parts and syntactically structured; (2) their simplest parts refer or denote things and properties in the world; (3) their meanings as wholes are determined by the semantical properties of their basic parts together with the grammatical rules that have generated the overall syntactic structures; (4) they have truth conditions …; (5) they bear logical relations of entailment or implication to each other." (Lycan 2003: 189) What is characteristic for the language of thought is not only that its parts represent, but also that it consists of sentence-like pieces that, due to their compositionality, have systematicity and productivity, as do natural languages (we can think a virtually unlimited number of thoughts and which thoughts one can think is connected in a systematic way).

### 2.1 Origin of Meaning?

This brings us to the problem. How is it possible that these symbols of a computational system have meaning? Fodor himself appears to see that this is problematic, at least sometimes: "How could a process which, like computation, merely *transforms one symbol into another* guarantee the causal relations between symbols and the world upon which … the meanings of symbols depend?" (Fodor 1994: 12-13). There seem to be two ways in principle: meaning is built-in (innate) or meaning is acquired. What I am trying to show here is that if LOCO is assumed, it cannot be acquired, leaving the option of built-in (innate) meaning.

### 2.2 A Short Line

The situation invites a very short line indeed: If language of thought computationalism is the manipulation of meaningful symbols, then the functioning of the language of thought (the "cognition" or the "thinking") *presupposes* the existence of meaningful symbols in the system. In other words, the system must have meaningful symbols *before* the language of thought can function. The acquisition of these meaningful symbols can thus not be the work of a language of thought.

So, if a newborn child's cognition is within language of thought, then a child must be born with meaningful symbols: language of thought computationalism presupposes meaningful symbols. Fodor himself has been supporting the idea of innate meaning for some time, of course, but many in the field want the language of thought computationalism without the nativism. Nativism is typically taken as optional but as the 'short line' shows, it is not. This "short line" is a simple argument against language of thought computationalism without semantic nativism—an argument we lack so far as far as I can see (see Fodor 2000; 2005; Pinker 2005a; 2005b).

## 3. Computing with Meaningless Symbols: Syntactic Computationalism

Given the problem described in the above "short line", it may be plausible to revert to a more modest version of computationalism: Mental computation is (or could be) purely syntactic. Of course, this does not exclude that the symbols could be interpreted by some observer; it just says that they have no meaning *for the system*. At first glance, this is what is the case with any conventional digital computing machine: For example, the operation of a set of switches that constitute an XOR-gate could be interpreted as be doing a logical operation, or as computing an addition, or as doing various other things. (The logical gates for exclusive or are the same as those for binary addition plus a "carrying over" of surpluses to the next digit.) The switches have no meaning for the system itself. When my pocket calculator displays the output "844$" or my washing machine displays "End", this means something to me, but not to the computer.

### 3.1 Symbol System, Technically

In order to understand the proposal of computationalism without presupposing meaning in the system, it is useful to gain a deeper understanding of what a computer, really does. The main characteristic of a digital computer is that is algorithmic. Any calculator can "carry out" a particular algorithm (and mechanical calculators were already constructed in the 17$^{th}$ Century by Schickard, Pascal and Leibniz). A computer, however, is programmable, that is which algorithm it carries out can be changed. The universal Turing machine is a model for a computer that can run any program, essentially by giving numbers to all the other simple Turing machines that can compute only one algorithm.

To understand computation, it is important to see that we can describe a computer on three (plus) levels of description:

*Physical level:* Some physical objects such as toothed wheels, holes in cards, states of switches, states of transistors, states of neurons, etc. are causally connected with each other—such that a state of one object can alter the state of another.

*Syntactical level:* The physical objects are taken to be tokens of a type (e.g. charge/no charge) and are manipulated according to algorithms. These algorithms are also stored and changed in the computer via some set of physical tokens (typically the same set). The manipulation follows the algorithms and only concerns these tokens as tokens, not their physical realization or their interpretation; it is "purely syntactical." To do this, the computer needs to recognize each token as of a type, as a basic symbol for this system, e.g. a 0 or 1 at the basic level of a binary system.

What I call the syntactical level could also be called the "form" of a computational procedure. This way of talking is aptly criticized by (Kuczynski 2006), who claims that there is no computational form without a semantics to identify tokens of types. In his discussion of physical form (morphology) vs. syntactical form Kuczynski fails to invoke levels of description and thus comes to the conclusion that there really is no such thing as a purely formal procedure distinguished by physical form alone (especially for logical inference).

Horowitz makes the related proposal that we need "computational externalism without relying on semantic externalism" (Horowitz 2007: 76). I argue elsewhere that we do not need the notion of semantics to solve what I call the "individuation problem", namely to explain what makes something to be a token of a computational type (see (Müller 2008a)).

*Symbolic level:* The physical objects that are manipulated on the logical level are taken to represent; they are (parts of) letters, numbers, words, images, vectors, concepts, … One could thus have one algorithm (on the syntactical level) that carries out several functions (on the symbolic level).

Piccinini's terminology, who discusses the problem of how to "individuate computational states" (Piccinini 2008) cuts across my position here: I do not adhere to a semantic view (since I allow description levels below the "symbolic" level), but neither do I subscribe to his view that that a computational state must be individuated functionally, in terms of function for a whole organism. I tend to think that this would pick out *one* level of description within my "symbolic" level: Piccinini's explanatory aim is different from mine. (For further discussion, see (Müller 2008a).)

I propose to have "3+" levels rather than "3" because each of the symbols on the symbolic level can symbolize something else in turn. Accordingly, one might distinguish several further levels within the symbolic level when describing a computational cognitive system, for example, the distinction between nonconceptual content and conceptual content, or the distinction between symbols and concepts (for the latter, see Gärdenfors 2000, ch. 7).

If we now describe a conventional van Neumann machine, e.g. a PC, at its syntactical level, rather than at its physical (realization) level, we will see basic operations on bits of main memory such as *read* (is this bit on or off?) and *write* (to this bit). These operations are combined by building in logical (Boolean) switches where one bit takes a particular state, given the state of two other bits. With the help of such switches, one can construct algorithms of switching patterns that perform particular tasks on the symbolic level, e.g. compare, add, … The computing process is a long sequence of such basic operations resulting in a memory state. Note that it is irrelevant for the syntactical description of the computer *how* a particular operation is carried out—one way to see this is to conceive of the computer as operating a Turing machine (see e.g. Davies 2000: 167).

### 3.2 Is there Computing Without Meaning?

After this initial clarification we can return to the proposal of syntactic computationalism. Some have claimed that this is per se impossible, that there could not really be a computing system, without any meaningful symbols. One prominent objection is that the system must be able to carry out *programs*, programs that are themselves encoded in symbols, and typically stored in memory. Does this not require following rules and understanding at some level? For example, in Searle's famous computation in the "Chinese Room" (cf. Preston and Bishop 2002; Searle 1980), Searle sits in the room and manipulates Chinese symbols according to manipulation instructions given in English: A language that he understands!

John Haugeland claims that in any computing system there are *primitive operations* of which the system knows how to carry them out (Haugeland 1985: 66). (This understanding may be prompted by the metaphorical use of 'command' and similar expressions at several levels of computer use.) Indeed, he says these must involve meaning: "The *only* way that we can *make sense of* a computer *as executing a program* is by understanding its processor as responding to the program descriptions as meaningful." (Haugeland 2002: 385) (cf. Boden 1990; 2006: 1414ff).

If this was right, in any computing system we would be back at our original problem: If there are "meaningful primitives" in any computing machine—where do they get their meaning? Our 'short line' would show that computationalism implies semantic nativism, generally. Or rather, semantic nativism must be true for any computer, given that we have working computers. All our computers would already have meaningful symbols built in!

I think it will be apparent form the discussion of descriptive levels above, that purely syntactic machines are in fact possible, however. We just need to be more careful when we say that the system "follows rules", or "executes programs." Wittgenstein famously distinguished between *following* a rule and *acting according to* a rule—and only the former requires that one understands the rule (gives it an interpretation). The computer does *not* literally follow a rule. Being in a particular state, given a particular input, it will perform a series of steps (e.g. switches) and produce a particular output, a memory state. The same happens when it is programmed, i.e. its switches are set (this even happens in the same central memory, in the case of a 'stored program' von Neumann machine). This is a purely causal, mechanical procedure that requires no understanding of a rule. It is no different from a can vending machine taking a particular input (my coins and my pressing a button), processing, and producing a particular output (the can).

The computing machine is just constructed in such a way that it will mechanically do what *we* call "carrying out a program", on the logical or even the symbolic level. We can describe the computer as "following a rule" and some of its states as "symbols" but that is entirely irrelevant to its functioning. A computer can be described on the symbolic level, but it must not have such a level. It may also, to repeat, be described differently on the symbolic level. The widespread resistance to calling computing "purely syntactical" (e.g. Davies 2000: 204-05; Hauser 2002; Preston 2002: 40-41; Rey 2002) perhaps due to the fact that this process is, of course, causal. It is not so much a formal procedure, but rather the syntactical properties of a physical procedure. On the syntactical level, one can say that the computer operates on meaningless symbols with programs that are meaningless to it.

Accordingly, the solution to symbol grounding cannot be to give basic rules, as does for example Hofstadter in his discussion of the matter. For his MU and MIU systems you assume that rules have meaning (Hofstadter 1979, chs. I & II, pp. 170, 264). If you do not, then you have to postulate that "absolute meaning" comes about somehow by itself, in "strange loops" (ch. VI and passim).

### 3.3 Can Purely Syntactic Computing Acquire Meaning? —A Challenge (the Longer Line)

So, how does syntactic computationalism, thus understood, fare with our problem of symbol grounding? The problem for a computationalist is that she has to construct a causal chain that does not involve any mental process at any stage that is other than purely syntactic. Meaning-involving processes such as attention, object tracking, object-files, interest, intention, etc. are not permitted.

Let us look at some lessons from history to understand the difficulty: I take the discussion about the so-called "causal theory of reference", originally developed by Putnam and Kripke in the early 1970ies, to have shown two things:

A) We want to grant causal connections between tokens of some kinds of symbols and their reference a role in the determination of the meaning of the symbols—in particular, we want to do this in the case of natural kind terms, such as "gold", where the stuff they refer to, the element *gold*, plays a role in the determination of what counts as gold and what does not. This is what Putnam called the "contribution of the environment." I say, "We want to grant" because it is important to see that Putnam's and Kripke's discoveries are discoveries about our linguistic intuitions.

B) The causal relations between, for example, the tokens of the word "gold" and the element *gold* are immensely complex and it is extremely hard to figure out the particular causal relation that should connect a particular token to its referent. A given token stands in any number of causal relations and none of these by itself distinguishes itself as the right one (for example, "gold" does not refer to jewelers shops or to chemistry textbooks or to metal or to undiscovered fake gold). What we need is a notion of "explanatory cause", the cause that is relevant for our explanatory intentions.

What is relevant here is not so much semantic externalism (that has lead to externalism about mental states) but Putnam's later critique of his own earlier causal theories of reference. This critique shows that a successful story of the causal relation between my tokens of "gold" and gold has to involve my *desire to refer* to that particular metal with that particular word. Putnam has tried to show this in his model-theoretic argument (Putnam 1981a: 34 etc.) and in the point that we need to single out what we mean by "cause," given that any event has several causes—whereas we need the one "explanatory" cause (Müller 1999; Putnam 1981b; 1985). This is supported by Wittgensteinian arguments to the effect that deixis is necessarily ambiguous (sometimes called the "disjunction problem"). When Kripke pointed at the cat (and Quine's native pointed at the rabbit), were they pointing at a cat, a feline, an animal, a flea, a color, or a symbol? When Putnam pointed at water, how much H2O did we need in the sample for reference to be successful?

The Putnam/Kripke story shows that the causal relation of a *linguistic* symbol to its referent must involve the *intention of speakers* to refer to a specific object or kind: otherwise it is underdetermined due to the multiplicity of causal chains.

Fodor himself seems to see an issue when he argues against language acquisition by non-linguistic thinkers as follows: "Plausibly, for example, learning English requires learning that the form of words 'it's raining' is properly used to communicate the thought that it's raining. How do you learn that sort of thing if you have the kind of mind that can't, even in principle, think about thoughts?" (Fodor 2003)

So, the problem is, how can a system acquire meaningful symbols without making use of cognition? Could there be a theory of language acquisition (or machine learning) that assumes a language can be learned by a system that has no cognitive processes? I propose that to develop such a theory is more than just a challenge: it cannot be done.

### 3.4 Relation to Searle's "Chinese Room Argument"

The same point can be illustrated in the terms used in Searle's "Chinese room argument" (cf. Preston and Bishop 2002; Searle 1980). Searle's central notion is "understanding" (of Chinese and of stories) and he claims, 1) that the symbol manipulator in the Chinese room should not be said to *understand* Chinese by virtue of his handling the symbols correctly and thus producing correct output, also that he has no chance of learning Chinese [both of this everybody agrees with], 2) that the whole system containing the Chinese room, with manipulation manuals and all, cannot be said to understand Chinese [the "systems reply"], not even if "sensory organs" (cameras, microphones, etc.) are added [the "robot reply"], since these supply "just more Chinese." He sometimes expresses this as saying that the system has syntax but no semantics for its symbols: that symbols in a system cannot acquire meaning due to mere symbol manipulation.

As several people have pointed out, 2) does not follow from 1). This does not mean that his argument fails, however. The upshot of the argument is, in my view, that Searle sets the task to explain *how* a system can understand Chinese given that the central symbol manipulator does not. After the Chinese Room Argument the belief that a symbol manipulating system can "understand" is in doubt and would require positive support.

Searle's claim is that he cannot learn Chinese by manipulating the symbols in his room, even if he tries hard—and then he expands this point to the whole system. But he already grants too much: Searle in the room *does understand* the symbols in the instructions for manipulation, *wants* to learn Chinese, *knows* that Chinese is a language, that some of its symbols refer and which world they refer to. None of these is given in an actual purely syntactic computational system. Given that there is literally *no* understanding, desire and knowledge in the actual Chinese Room of a syntactic system (there are no intentional states), there is even less reason to believe that there is in the whole system.

The argument presented above thus goes some way towards closing the gap in Searle's argument by explaining why symbol manipulation, even under causal interaction with the environment, cannot produce intention. The system will not acquire meaningful symbols because it lacks everything necessary, specifically it has no *desire* to do so (it has no desires directed at anything). The situation is thus worse than in Searle's "Chinese Room", where Searle tries to show that an intelligent agent operating a purely syntactical system *cannot* acquire meaning. We only need to claim that a purely syntactical system itself *will not* acquire meaning—even if it could. (For the opposing view in 'epigenetic robotics', see, for example (Steels 2008) and cf. (Taddeo and Floridi 2005).)

On a cautionary note, just like Searle, we do not claim to have found any bounds as to what can be done with purely syntactic computing. Clearly, advanced AI systems (and perhaps "lower" animals) have achieved impressive feats without the "meaningful symbols" we have been asking for and which humans surely possess (Müller 2007).

This look into the Chinese room might leave a paradoxical air; one might wonder what that magical bit is which allows humans and other animals what computers cannot have. My suggestion here is that this bit has to be something that is not computational—and I think *desire* is a good candidate.

## 4. Taking Stock

### 4.1 Some Conclusions

What we have seen so far is that:
   1) A language of thought computational system presupposes innate meaning,
   2) A purely syntactical computational system is possible,
   3) A purely syntactical computational system could only acquire meaning if that process does not involve any mental states with intention (e.g. desires, beliefs, attention, …).

What we have not seen is whether there is another version of computationalism that could save the day. Perhaps there is computation without symbols or there is information processing in ways other than computing? Let us take a brief look at the options.

### 4.2 Vacuous Computationalism

Searle has repeatedly said that whether a system is a computer or not depends on its interpretation by some observer, a syntactic property is an observer-relative notion This is why he comes to the prima facie surprising conclusion that "The brain is a computer, in the sense that it instantiates computer programs..." because "everything is a digital computer at some level of description" (cf. Preston 2002: 42-44; Searle 2002: 224).

Whether this view is true or not (I tend to think it is not (cf Piccinini 2007)), as Searle knows, this makes computationalism vacuous. Clearly, computationalism cannot be the claim that, if an observer likes to see it that way, the brain is a computer, and so is a train, a tree or a bumblebee.

### 4.3 Non-Symbolic Computing and "Information Processing"

There are cognitive scientists that use the word "computational" in a much weaker sense than the one defined above—in fact, the plethora of definitions is depressing: I counted 9 different ones, most of which are obviously either too narrow or too broad, in a recent exchange between Pinker and Fodor (Fodor 2005; Pinker 2005a; 2005b):
   1) Literally being a Turing machine with tape and all (attributed to Fodor by Pinker 2005b: 6).
   2) "Cognitive architecture is Classical Turing architecture" (Pinker 2005b: 6).
   3) Having "the architecture of a Turing machine or some other serial, discrete, local processor" (attributed to Fodor by Pinker 2005b: 22). False attribution, since in 2000, Fodor did not mention the possibility of *other* processors. Suggests that

    "architecture" means physical setup (tape and reader), after all—see problems in 2).

4) Being 'Turing-equivalent', in the sense of 'input-output equivalent' (Fodor 2000: 30, 33, 105n3).
5) Being 'defined on syntactically structured mental representations that are much like sentences' (Fodor 2000: 4).
6) Being supervenient "on some syntactic fact or other"—"minimal CTM" (Fodor 2000: 29).
7) Being "causally sensitive to, and only to, the *syntax* of the mental representations they are defined over" [not to meaning] AND being "sensitive only to the *local* syntactic properties of mental representations" (upshot in Fodor 2005: 26).
8) "In this conception, a computational system is one in which knowledge and goals are represented as patterns in bits of matter ('representations'). The system is designed in such a way that one representation causes another to come into existence; *and* these changes mirror the laws of some normatively valid system like logic, statistics, or laws of cause and effect in the world." (Pinker 2005b: 2).
9) "…human cognition is like *some kind* of computer, presumably one that engages in parallel, analog computation as well as the discrete serial variety" (Pinker 2005b: 34).

    One prominent idea is that computing is somehow "information processing." (cf. Müller 2008b) But information processing could take many forms, some of which are not computational. There are many systems that could be used to compute but should not be called a computer. Dynamical systems in the sense of van Gelder (van Gelder 1995) are one example. Another are analogue systems, such as slide rules, mechanical (non-digital) adding machines, scales, tubes, etc. So, even if computing is information processing, what distinguishes it from *other* forms of information processing—some of which may even produce the same results? Surely this must be the *mechanism* by which it achieves that processing: namely computation (i.e. performing algorithms). There are at least two notions of algorithm possible here, depending on whether the step-by-step process is one of symbol manipulation or not (for a discussion, see Shagrir 1997). (Harel 2000).

    I would therefore make the terminological suggestion to distinguish between "computationalism" and "information processing" as paradigms for cognitive science.

### 4.4 Outlook: Analogue and Hybrid systems

The pages above present a reason to believe that the mind is not a computational system, unless it has semantics built in. However, there is still good reason to think that some parts of the human mind are computational, even if the problems explained show that it is not *only* that. Perhaps the picture that emerges is that of a hybrid and modular mind where some modules are computational but many are not (Fodor 2000: 99 etc.). Some of the non-computational systems will be mathematically describable, perhaps as dynamic systems, and can thus be simulated on digital computers to some degree of accuracy. Perhaps some of these cognitive processes are intentional but neither syntactic computation nor LOT computation, and together with the embodi-ment of the whole, they can explain how meaning can be acquired (if it is not built in).

**Acknowledgments.** My thanks to people with whom I have discussed this paper, especially to Thanos Raftopoulos, Kostas Pagondiotis and the attendants of the "Philosophy on the Hill" colloquium. I am very grateful to two anonymous reviewers for detailed written comments.